\documentclass[12pt]{spieman}  % 12pt font required by SPIE;
\usepackage{amsmath,amsfonts,amssymb}
\usepackage{graphicx}
\usepackage{setspace}
\usepackage{tocloft}
\usepackage{graphicx}
\graphicspath{{figures/}}
\usepackage{amsmath}
\usepackage{amssymb}
\usepackage{booktabs}
\usepackage{multirow}
\usepackage[table,xcdraw]{xcolor}
\usepackage{hyperref}
\usepackage{lineno}
\usepackage{soul}
\usepackage{color}

\title{UMFA: A photorealistic style transfer method based on U-Net and multi-layer feature aggregation}

\author[a]{Dongyu~Rao}
\author[a,*]{Xiao-Jun Wu}
\author[a]{Hui Li}
\author[b]{Josef Kittler}
\author[b]{Tianyang Xu}
\affil[a]{Jiangnan University, Jiangsu Provincial Engineerinig Laboratory of Pattern Recognition and Computational Intelligence, School of Artificial Intelligence and Computer Science, Lihu Avenue, Wuxi, China, 214122}
\affil[b]{University of Surrey, Centre for Vision, Speech and Signal Processing, Guildford, UK., GU2 7XH}

\begin{document}

\maketitle
\linenumbers 

\begin{abstract}
In this paper, we propose a photorealistic style transfer network to emphasize the natural effect of photorealistic image stylization. 
In general, distortion of the image content and lacking of details are two typical issues in the style transfer field. 
To this end, we design a novel framework employing the U-Net structure to maintain the rich spatial clues, with a multi-layer feature aggregation (MFA) method to simultaneously provide the details obtained by the shallow layers in the stylization processing. 
In particular, an encoder based on the dense block and a decoder form a symmetrical structure of U-Net are jointly staked to realize an effective feature extraction and image reconstruction. 
Besides, a transfer module based on MFA and "adaptive instance normalization" (AdaIN) is inserted in the skip connection positions to achieve the stylization. 
Accordingly, the stylized image possesses the texture of a real photo and preserves rich content details without introducing any mask or post-processing steps. 
The experimental results on public datasets demonstrate that our method achieves a more faithful structural similarity with a lower style loss, reflecting the effectiveness and merit of our approach. 
\end{abstract}

% Include a list of up to six keywords after the abstract

% Include email contact information for corresponding author
{\noindent \footnotesize\textbf{*}Xiao-Jun Wu,  \linkable{wu$_-$xiaojun@jiangnan.edu.cn} }

\begin{spacing}{2}   % use double spacing for rest of manuscript

\section{Introduction}
\label{sect:intro}  % \label{} allows reference to this section
Photorealistic image stylization aims at realizing a style transfer between the two real photos.
The typical styles to be transferred include season \cite{laffont2014transient}, weather and illumination \textit{etc.} \cite{luan2017deep}. 
It can be regarded as a sub-problem of the wider class of style transfers. 
It differs from the general style transfer task which employs realistic and artistic images to obtain artistic content images. 
The task of photorealistic style transfer is to achieve a realistic photo texture with natural stylized output. 
Therefore, the photorealistic image stylization has a wide range of applications and excellent development prospects in photo generation, picture processing, \textit{etc.}.

In the field of computer vision, style transfer is usually studied as a generalization problem of texture synthesis \cite{akl2018survey}, \textit{i.e.}, extracting the texture from the source and transmitting it to the target \cite{efros2001image}. 
Traditional methods mainly focus on model building and texture synthesis \cite{song2016sufficient}. 
However, traditional methods can only extract shallow features of the image without involving deep features, resulting in the poor quality of the final appearance of the style transfer for complex images. 
Besides, the constructed model is effective for a predefined specific style only.
 
With the emergence of deep learning, the use of deep features extracted by neural networks for transforming a style has been widely developed \cite{gatys2016image,zhang2018real, zhang2018multi}. 
Gatys \textit{et al.} \cite{gatys2016image} proposed one of the first style transfer methods based on deep convolutional neural networks \cite{krizhevsky2012imagenet}. 
This method uses a convolutional neural network to extract deep features and exploits the Gram matrix to characterize the style in the image. 
By iteratively updating the reconstructed image, it realizes the stylization of the input content image. 
This method can achieve a transfer of style effects even between complex images. 
In addition, it solves the problem of applying a single model to different styles. 
This seminal work has opened up a new field of research named Neural Style Transfer (NST) \cite{jing2019neural}. 
However, given a new image, each process of the stylization model has to be retrained, increasing the computation. 
Moreover, the tendency to distort the content image is notable. 
Subsequently, methods to improve the quality of stylization and to accelerate the speed have been proposed \cite{gatys2017controlling,gu2018arbitrary,cheng2019structure, hu2020aesthetic}. 
Although these methods achieve impressive artistic style transfer effects, they cannot suppress the distortion and deformation of the content. 

With the development of the neural style transfer field, in addition to the study of artistic style transfer, other aspects of style transfer techniques are also of interest, such as style transfer of high-resolution images, which is used to process HD images \cite{texler2020arbitrary,li2019high}, text style transfer that changes text attributes \cite{yang2020te141k,li2019chinese} and photorealistic style transfer between real images \cite{wang2020photographic,yoo2019photorealistic}. 
 
Traditional methods of photorealistic style transfer are mainly based on color and tone matching \cite{sunkavalli2010multi,pitie2005n,yoo2013local}. 
With the development of the neural style transfer, researchers have found that neural style transfer could be used to realize the task of photorealistic image conversion. 
Based on the algorithm proposed by Gatys, Fujun Luan \textit{et al.} \cite{luan2017deep} were the first to tackle the photorealistic style transfer task. 
Using the local affine transformation, inspired by the Matting Laplacian \cite{levin2007closed} in the color space (RGB), the distortion of the content in the generated image is significantly reduced. 
Subsequently, a series of algorithms, such as PhotoWCT \cite{li2018closed}, WCT$^{2}$ \cite{yoo2019photorealistic}, and other methods \cite{an2020ultrafast,wang2020photographic} were proposed with promising performance. 
They improved the quality of the generated images by changing the structure of the existing methods and introducing post-processing steps. 
However, the appearance of stylized images produced by these methods is not natural. 
In addition, these methods introduce more operating steps, increasing the processing time.

The key issue regarding the photorealistic image style transfer is how to keep the structure and details of the content image while introducing the target style to achieve a natural stylized effect. 
Meanwhile, there is one problem to be solved, \textit{i.e.}, realizing the photorealistic style transfer in real-time for the needs of real-world applications. 
Current methods cannot solve this problem very well.

The specific objective of this study is to remove the distortion of the content in the stylized image and to accelerate the natural stylization processing. 
In \cite{li2018closed,yoo2019photorealistic,huang2017arbitrary}, the VGG network is used to extract deep features to achieve image stylization. 
But these methods do not make full use of the contextual information. 
In contrast, we propose to use dense block-based U-Net for feature extraction and image reconstruction, retaining more image detail. 
Besides, feature aggregation is an effective feature enhancement method in computer vision tasks. 
In order to tightly connect the features, a multi-layer feature aggregation (MFA) method is proposed and used to enhance the representation ability of the extracted features. 
In summary, we propose a new photorealistic style transfer method named UMFA based on U-Net \cite{ronneberger2015u} and multi-layer feature aggregation \cite{yu2018deep}. 
The main contributions of our method UMFA are as follows:

(1) A dense neural network architecture unit is used in a U-Net based style transfer framework. The dense block enhances the quality of feature transferring. The benefit of the combination of dense block and U-Net is the retention of spatial detail and the ability to express image content.

(2) A multi-layer feature aggregation method is used to fuse multi-scale features. The MFA module strengthens the ability to express features, improving the stylization effect by introducing aggregate features in each layer.

(3) The stylization process is carried out in the skip connection. Through a cascade processing method, the directly stylized features are inserted into the image reconstruction process to achieve a natural stylization effect.

The rest of our paper is structured as follows. In Section \ref{sec2}, we briefly review the photorealistic style transfer method based on deep learning and the network structure used in this paper. In Section \ref{sec3}, we present the proposed UMFA method in detail. In Section \ref{sec4} we report, analyze, and compare the experimental results. Finally, we summarize the main achievements and draw the paper to the conclusion in Section \ref{sec5}.

\section{RELATED WORKS}
\label{sec2}
Recently, many researchers have adopted deep learning-based methods to realize photorealistic style transfer between images and achieved very promising results\cite{luan2017deep,li2018closed,yoo2019photorealistic}. In this section, first, we review several photorealistic image stylization methods based on convolutional neural networks. Then we introduce the computational process known as  the U-Net\cite{ronneberger2015u} structure. Finally, the problem of feature aggregation is highlighted in the context of automatic style transfer and the main feature aggregation  mechanisms are discussed \cite{an2020ultrafast}.
\subsection{Deep Learning-based Photorealistic Style Transfer Methods}
In 2016, Gatys et al. proposed the use of a VGG network\cite{vgg} based method  to extract features and to perform art style image transfer. The authors deployed  a pre-trained VGG model to obtain deep features of the input image and a style image, and showed that a stylized image with both the target content and the required style can be generated by optimizing a random noise image iteratively. At the same time, their method of expressing style information with Gram matrix has also been widely used. The Gram matrix is composed of the inner product of the two vectors, so the Gram matrix can reflect the internal relationship between the vectors in the group of vectors. The deep feature of an image can represent the structure, texture and other information of the image, so when the Gram of the deep features of two images are similar, they have similar internal connections, that is, they have similar styles.

Subsequently, Johnson et al.\cite{johnson2016perceptual} proposed an image style transfer method that uses a perceptual loss to iteratively generate an optimized model named  rapid style transfer. It greatly improves the speed of the algorithm, but the model can  work only for one style.

In 2017, Huang et al.\cite{huang2017arbitrary} proposed an adaptive instance normalization (AdaIN) method inspired by the notion of instance normalization (IN)\cite{dumoulin2016learned,ulyanov2017improved} for transferring arbitrary image styles. 
From a statistical point of view, the style information of an image is also a kind of statistical information. When two images have the same distribution, they also have similar styles. The "adaptive instance normalization" is a method which changes the mean value and variance of the content image feature and keeps it consistent with the mean value and variance of the style image to realize style transfer.

Li et al.\cite{li2017universal} developed a whitening and color transform (WCT) method for style transfer based on a feature transform. These methods can achieve a real-time, arbitrary style transfer.

In 2018,  Li et al.\cite{li2018closed} proposed a technique called PhotoWCT, which further improves the baseline established in \cite{li2017universal}. This method changes the upsampling step to the unpooling operation in the decoder, to smooth the final  result. This method maintains the spatial consistency of the content image and the stylized image. In 2019, Yoo et al.\cite{yoo2019photorealistic} proposed WCT$^{2}$ method inspired by WCT. This method uses wavelet pooling instead of maximum pooling in \cite{li2018closed}. Wavelet pooling retains the image details, while the model introduces richer style elements through multi-level stylization operations.

The wide application of deep learning makes the design of the neural network structure an important factor affecting the model performance. Neural architecture search (NAS)\cite{nas} is a technology for automatically designing efficient neural network structures. In 2020, An et at.\cite{an2020ultrafast} designed a model for ultrafast photorealistic style transfer through neural architecture search. The basic structure of this method is similar to WCT$^2$\cite{yoo2019photorealistic}. They searched for a neural architecture performing a multi-layer stylization  to optimise speed and convergence.

In addition to stylization methods based on the encoder and decoder networks, generative adversarial networks (GAN) are also a hot research topic. Owing to their characteristics, the images generated by a GAN tend to have realistic  photo textures and natural appearance. Some methods , e.g. Pix2Pix\cite{isola2017image}, AutoGAN\cite{cao2020auto} and others\cite{zhou2019branchgan,zhu2017unpaired} have reported remarkably good style transfer results.

\subsection{The U-Net Architecture}
In 2015, Ronneberger et al.\cite{ronneberger2015u} proposed a network structure for medical image segmentation named U-Net. The structure consists of two main parts: encoder and decoder. The encoder includes convolutional layers and a pooling layer, which is constituted by a downsampling module. The decoder is a corresponding upsampling module. A skip connection between the corresponding layers is added to retain rich detailed image information. In addition to medical image segmentation, U-Net has achieved good results in semantic segmentation, image restoration, and enhancement\cite{liu2017image}. Thanks to the U-Net structure, the image context is preserved and  effectively used to reconstruct the input image.

\subsection{The Feature Aggregation}
With the emergence of deep learning, deep features play an important role in computer vision tasks. In order to improve the performance, the architecture of  deep neural networks is increasingly deeper and wider \cite{he2016identity,he2016deep} to enhance the modeling capacity and flexibility. Besides, tightening connections between layers is also an important factor affecting the performance of deep neural networks, as proved by the idea of skip connection\cite{huang2017densely}. Feature aggregation can closely connect features of different scales and levels to achieve a better quality of image reconstruction, as demonstrated in \cite{yu2018deep}. 

\begin{figure}
\begin{center}
\begin{tabular}{c}
\includegraphics[width=0.9\textwidth]{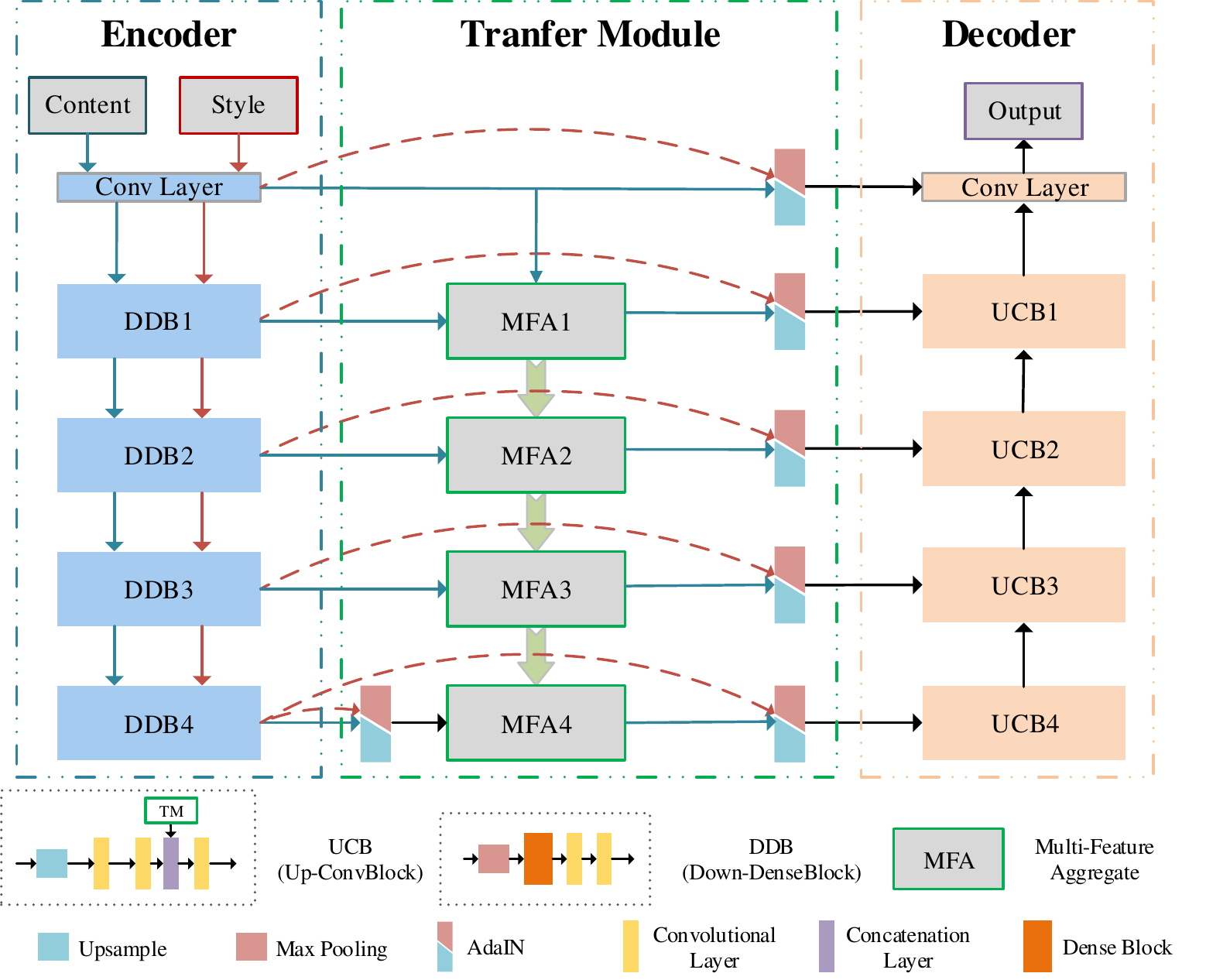}
\end{tabular}
\end{center}
\caption 
{ \label{fig1}
The framework of our method. } 
\end{figure}

Inspired by these works, we adopt the U-Net structure for the task of style transfer and enhance its modeling capability by endowing it  with a multi-layer feature aggregation mechanism. Finally, the merits of the proposed UMFA method based on U-Net and multi-layer feature aggregation are demonstrated by extensive experiments.

\section{PROPOSED PHOTOREALISTIC STYLE TRANSFER METHOD}
\label{sec3}
In this section, we introduce the proposed UMFA method. First, we present the UMFA framework in Section \ref{sec:network}. The details of its training are presented in Section \ref{sec:train}.

\subsection{UMFA: Photorealistic Style Transfer Network}
\label{sec:network}
As shown in Fig. \ref{fig1}, UMFA consists of three main parts: encoder, decoder, and a transfer module. The encoder and decoder constitute the basic symmetric structure of U-Net. %The transfer module performs the style transfer  by means of MFA blocks and AdaIN blocks. 
The encoder includes DDB blocks and a Conv Layer block. The decoder includes UCB blocks and a Conv Layer block. The remaining part of the architecture is the "transfer module" which includes MFA blocks and AdaIN blocks. In the encoder, the red and blue solid lines represent the features of the style image and content image, respectively. 

In the "transfer module", blue solid line represents the characteristics of the content image. To make it easier to distinguish in the figure, the red dashed line is used to represent the characteristics of the style image. The black lines indicate stylized features. The features of the style and content images are extracted by the encoder as a prerequisite to a style transfer through the "transfer module". The decoder reconstructs the output image.

\begin{figure}
\begin{tabular}{c}
\centering
			\includegraphics[width=0.9\textwidth]{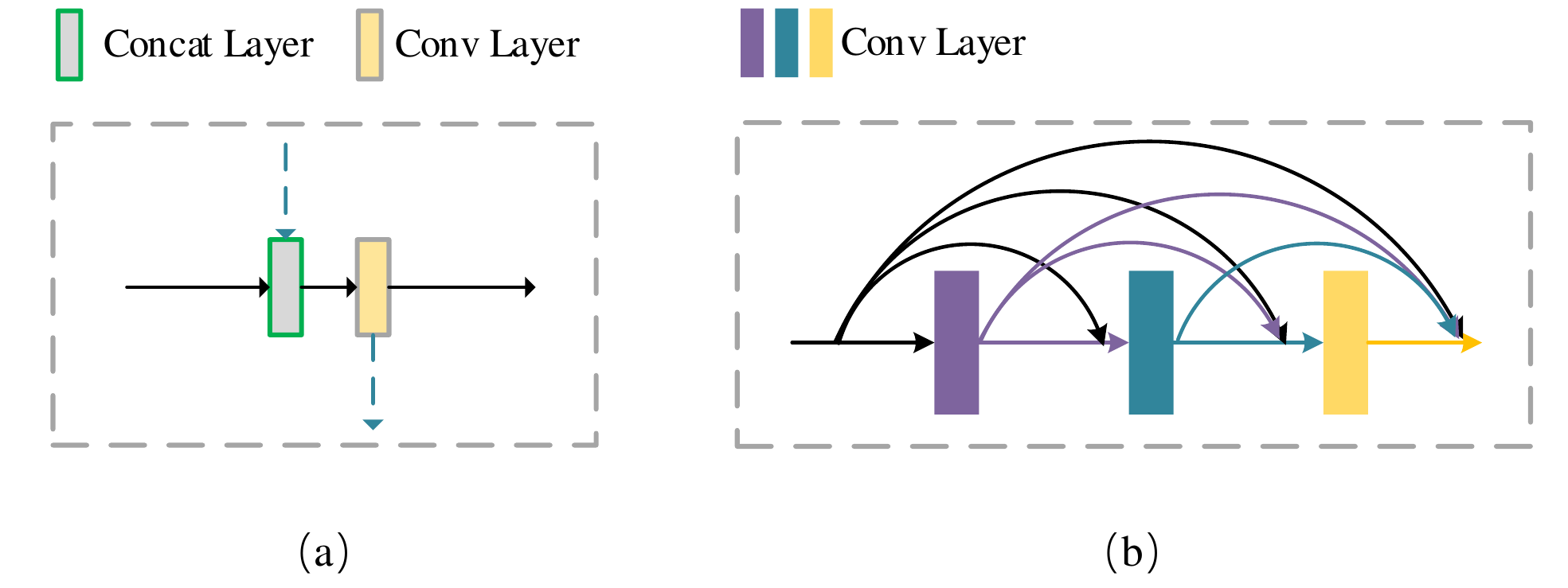}
\end{tabular}

	\caption 
	{ \label{fig2}
		(a)The framework of multi-layer feature aggregation (MFA);(b)The framework of Dense Block. } 
\end{figure}

In Fig. \ref{fig1}, "Conv Layer" represents convolutional layer. "DDB" means downsampling dense block\cite{li2018densefuse} which contains a max-pooling layer, a dense block, and two convolutional layers. The framework of dense block is shown in Fig. \ref{fig2} (b). Three convolutional layers are densely connected to form a dense block.
"UCB" denotes upsampling convolutional block which has an upsampling layer, three convolutional layers and a concatenation module that connects information coming from the "transfer module". "MFA" represents multi-layer feature aggregation at skip connection. In Fig.\ref{fig2} (a), the structure of "MFA" is displayed, where "Concat Layer" represents the cascade of the features passed from the previous with the features of layer. Then the AdaIN block uses "adaptive instance normalization" to transfer the style. The input images pass through a convolutional layer containing 3$\times$3 filters. The number of channels is increased to 32 without changing the feature size, which facilitates subsequent convolution and pooling operations. Each of the four downsampling blocks consists of a max-pooling layer, a dense block and two convolutional layers. Among these, two convolutional layers with 1$\times$1 filters are used to change the feature dimension. The max-pooling layer implements downsampling operations to halve the feature size. Thus, with each pass through a downsampling block, the width and height of the feature map becomes half of the original size and the number of channels is doubled. In this way, we can obtain multi-scale image information and enhance the model's ability to process high-resolution images.

The decoder mirrors the encoder structure, which is composed of 4 upsampling blocks to form a U-Net structure. The upsampling block consists of an upsampling layer, three convolutional layers and a concatenation module that receives the information coming from the "transfer module". The concatenation module combines the upsampled feature and the corresponding encoder feature received from the "transfer module". Using the convolutional layer to restore the required number of channels, we  enhance the multiplexing of features and recover more details of the input image.

After extracting multi-scale features, the "transfer module" is used to process the features of the content and style images with the aim of style transfer. We introduce a feature aggregation mechanism\cite{an2020ultrafast} in the "transfer module" to connect the multi-scale features of each layer. Feature aggregation enables the network to integrate spatial information from different scales and retain image detail. In order to keep more detailed information in each layer, we propose a multi-feature aggregation (MFA). As shown in Fig. \ref{fig2}, we insert the features of the previous to the next layer, and repeat the same feature aggregation process at the subsequent layers. Then the AdaIN method is used to stylize these aggregated features in each layer.

\begin{figure}
\begin{tabular}{c}
\centering
\includegraphics[width=0.9\textwidth]{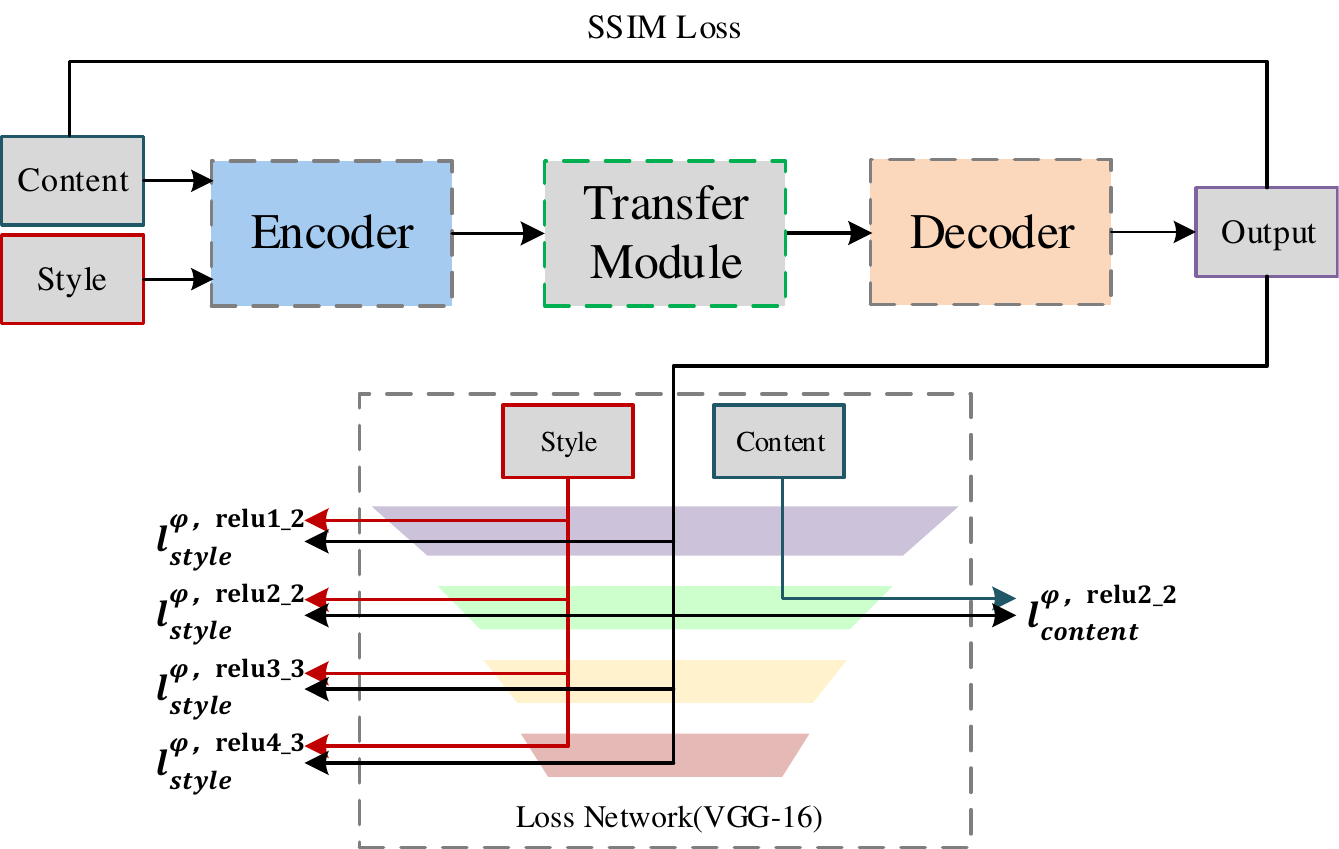}
\end{tabular}
	\caption 
	{ \label{fig3}
		The framework of the training process of UMFA. } 
\end{figure}

\subsection{Trainig Phase}
\label{sec:train}
In the training phase, we train the network end-to-end to learn a photorealistic style transfer model. The network framework is shown in Fig. \ref{fig3}. In order to better extract and use image features, we use a pre-trained VGG-16 network. The trapezoids of different colors in the loss network represent different blocks. Here we use the features extracted from the first four blocks. The $relui_-j$ indicates the output in loss network, that $i$ indicates which block it is in VGG and $j$ indicates which layer is in the block.

In the training phase, the loss function $L_{total}$ given in Eq.\ref{total} is defined as
\begin{equation}\label{total}
L_{total}=\alpha L_{style} + \beta L_{content} + \gamma L_{SSIM}  
\end{equation}
where $L_{content}$ and $L_{SSIM}$ respectively are content loss and structure similarity(SSIM) loss between the content image and output image. $L_{style}$ is style loss between the style image and the output image.  $L_{content}$ and  $L_{style}$ relate  to the perceptual loss\cite{johnson2016perceptual}. $\alpha$, $\beta$ and $\gamma$ are the weights of the above three losses. $O$, $C$ and $S$ represent the output image, content image and style image, respectively.

$l^{\varphi,j}_{content}$ is defined as the L2 norm squared of the feature which is extracted by ReLU layer in the loss network $\varphi$:
\begin{equation}\label{eq-con}
l^{\varphi,j}_{content}(O,C) = \frac{1}{C_jH_jW_j}||\varphi_j(O)-\varphi_j(C)||^{2}_2
\end{equation}
\begin{equation}\label{eq-con2}
L_{content}(O,C) = l^{\varphi,relu2_-2}_{content}(O,C)
\end{equation}
Different from the perceptual loss, we choose the $relu2_-2$ as the feature layer  for the content loss.

On the basis of the former, Gram matrix of the feature is used to represent the style feature. $l^{\varphi,j}_{style}$ is calculated by Eq. \ref{eq-sty} 
\begin{equation}\label{eq-sty}
l^{\varphi,j}_{style}(O,S) = \frac{1}{C_jH_jW_j}||G(\varphi_j(O))-G(\varphi_j(S))||^{2}_2
\end{equation}

\begin{equation}\label{eq-sty2}
L_{style}(O,S) = \sum_{j}^{relu1_-2\cdots relu4_-3} l^{\varphi,j}_{style}(O,S)
\end{equation}
where $G(\cdot)$ denotes the calculation of Gram matrix.$C, H$ and $W$ are the number of channels, height and width of the j-th feature map(64*256*256, 128*128*128, 256*64*64, 512*32*32). Finally, the style loss 
$L_{style}$ is the sum of all outputs from the four ReLU layers.

The SSIM loss $L_{style}$ is calculated by Eq. \ref{eq-ssim}
\begin{equation}\label{eq-ssim}
L_{SSIM} = 1 - SSIM(O,C)
\end{equation}

where SSIM($\cdot$) represents a structure similarity measure\cite{wang2004image}. Small values of $L_{SSIM}$ and $L_{content}$ means the output image $O$ and the content image $C$ are similar. A small  value of $L_{style}$ signifies that the output image and the style image are similar.

\section{EXPERIMENTAL RESULTS AND ANALYSIS}
\label{sec4}
In this section, we introduce the experimental settings used during the testing phase. Then we report the results of an ablation study. Finally, we present a subjective evaluation and some objective evaluation indicators to compare the proposed method with the state of the art.

\subsection{Experimental Setting}

For photorealistic style transfer, any two photoes can be used as style and content images. In the training, we randomly divide the 80,000 pictures from MS-COCO\cite{lin2014microsoft} into two equal parts as the data set, one part is the style image, the other part is the content image. The training process includes two epochs. The training image size is set to 256$\times$256 pixels. The Adam optimizer is used for training. The learning rate is 0.0001 and unchanged. Our model is implemented with NVIDIA TITAN Xp and Pytorch.

In the test,  we use the image pairs provided in \cite{luan2017deep} as data set. And we verified it on other more test sets\cite{yoo2019photorealistic}. Each set includes content, style. No masking or post-processing is used in the evaluation in order to verify the effectiveness of the methods.

We choose three typical methods for comparison, including: "adaptive instance normalization" (AdaIN)\cite{huang2017arbitrary}, "photorealistic image stylization based on the whitening and coloring transform" (PhotoWCT)\cite{li2018closed} and "photorealistic style transfer via wavelet transforms" (WCT$^2$)\cite{yoo2019photorealistic}. The results are obtained  using  publicly available codes for these methods and compared with UMFA on the same benchmark.

Two evaluation criteria are used to quantitatively compare UMFA with other methods: structural similarity (SSIM) and Gram loss. Structural similarity (SSIM) measures the similarity of the stylized image with the content image. The Gram loss is the mean square error between the stylized image Gram matrix and the style image Gram matrix that represents the similarity of the style. Larger SSIM and smaller Gram loss mean better performance.

\subsection{Ablation Study}

\begin{figure}
\begin{tabular}{c}
\centering
			\includegraphics[width=0.9\textwidth]{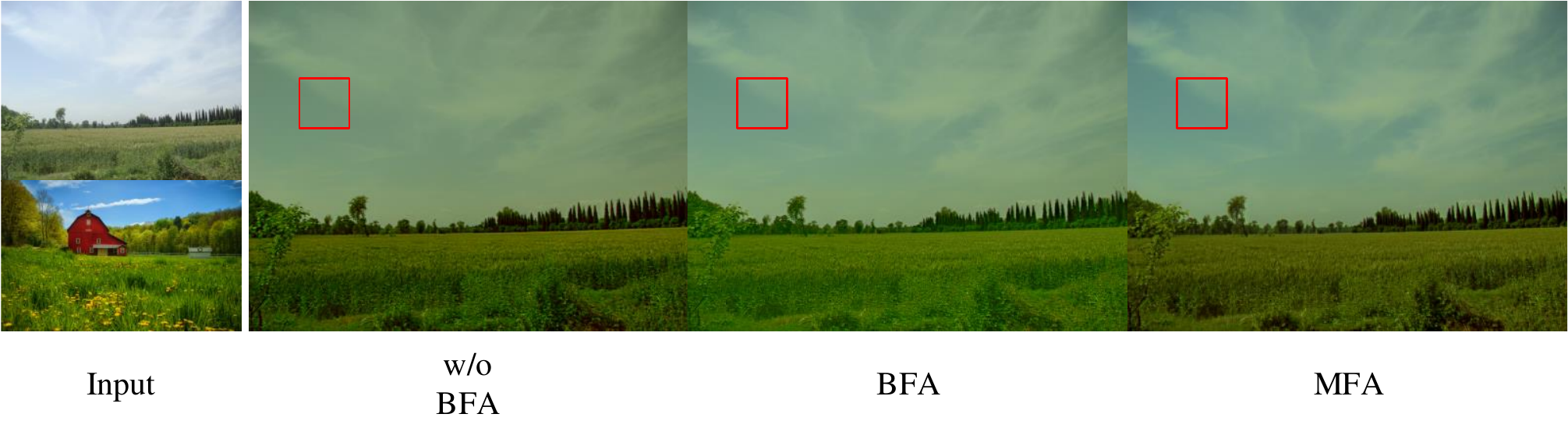}
\end{tabular}

	\caption 
	{ \label{fig4}
		The photorealistic stylization empowered by feature aggregation. (input pair (top: content, bottom: style)) } 
\end{figure}

\textit{1)\textbf{ Feature aggregation strategy:}} Feature aggregation can enrich the details of stylized images. Inspired by the bottom feature aggregation method (BFA)\cite{an2020ultrafast}, we propose a multi-layer feature aggregation (MFA). In order to verify the effectiveness of the multi-layer feature aggregation structure, we experiment with three models: no aggregation, BFA and MFA. At this 
time, other parameters are set as follows: $\alpha$=0.8, $\beta$:$\gamma$ is 1:1 ($\beta$=1, $\gamma$=1). In Fig. \ref{fig4}, we show a set of images to compare the results obtained with different aggregaation strategy. The plots of the total loss and style loss are shown in the Fig. \ref{fig8}. The results of the three aggregaation strategies are summarized in Table \ref{table3}.

\begin{figure}
\begin{tabular}{c}
\centering
			\includegraphics[width=1\textwidth]{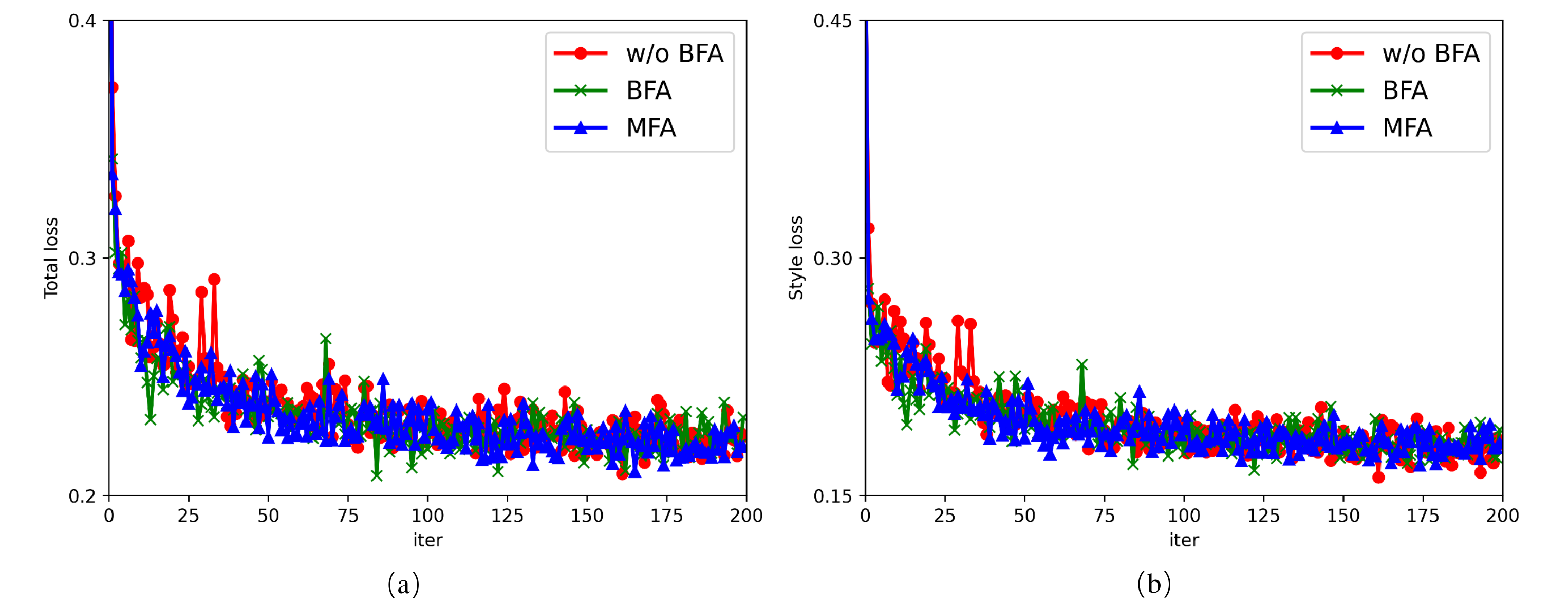}
\end{tabular}
	\caption 
	{ \label{fig5}
		A plot of the evolution of the total loss(a) and style loss(b) during training (feature aggregation strategy). Each scale on the abscissa represents 100 iterations.} 
\end{figure}

As shown in Fig. \ref{fig4}. BFA and MFA method have similar performance, and their effect is better than inapplicable feature aggregation. The color of the sky in the red frame is closer to the style image. In Fig. \ref{fig5}, the convergence curves of the three methods are very close, so we operate on the values (the reciprocal of the negative logarithm). We can find that the blue line(MFA) fluctuates less during convergence which means that its convergence process is more stable. In Table \ref{table3}, the results of the three methods have similar SSIM, but the result of the MFA method has a smaller Gram loss which means the results of MFA method are closer to the style of the targets. Therefore, the MFA method is used as a feature aggregation strategy.

\begin{table}
	\centering
	\caption{\label{table3}The quantitative evaluation results achieved by  different feature aggregation strategies}
	\setlength{\tabcolsep}{6mm}{
\begin{tabular}{cccc}
			\toprule 
			& w/o BFA & BFA & MFA \\
			\midrule 
			Gram loss & 9.476 & 9.486 & \textbf{8.996} \\ 
			SSIM & 0.620 & \textbf{0.625} & 0.612 \\ 
			\bottomrule 
\end{tabular}} 
\end{table}

\begin{figure}
	\centering	
	\includegraphics[width=0.9\textwidth]{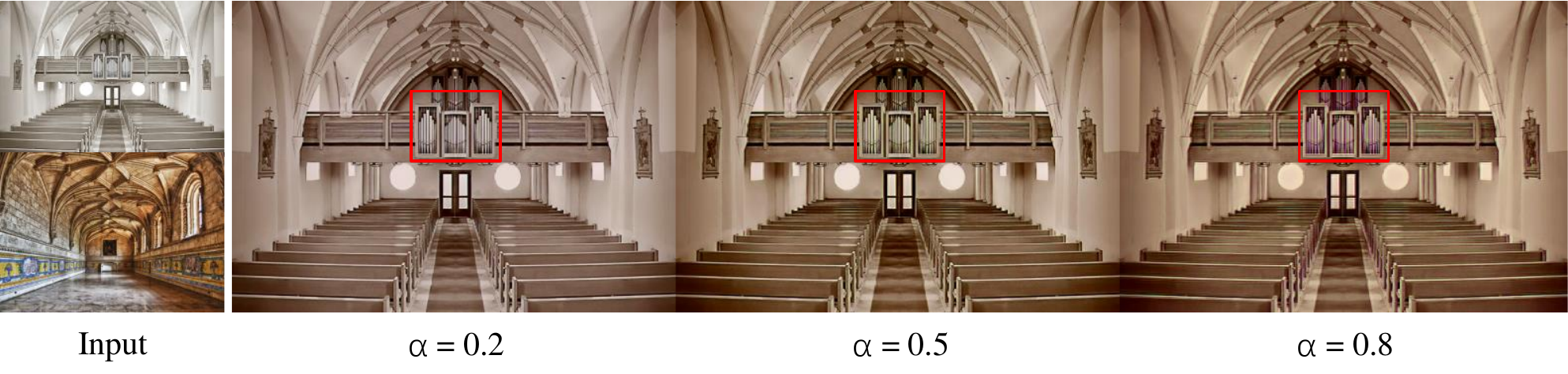}	
	\caption 
	{ \label{fig6}
		The photorealistic stylization results with different parameter $\alpha$. (input pair (top: content, bottom: style)) } 
\end{figure}

\begin{figure}[!t]
	\centering	
	\includegraphics[width=1\textwidth]{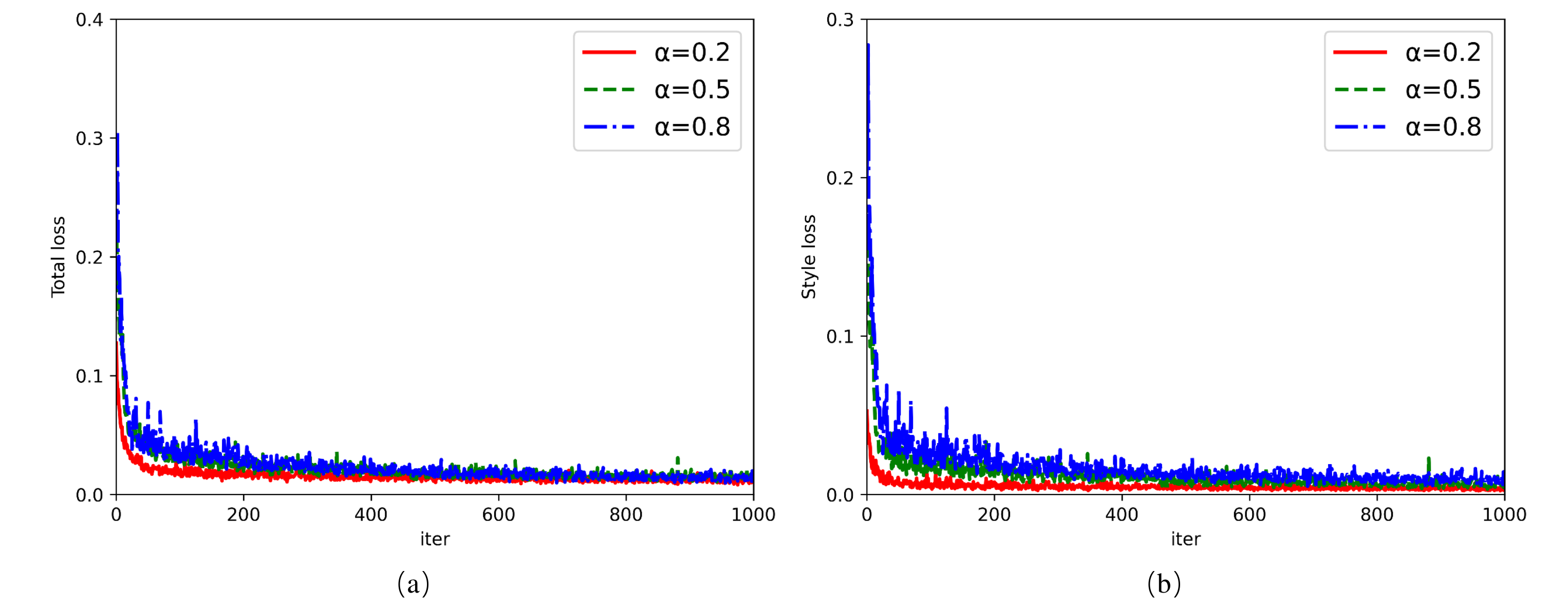}	
	\caption 
	{ \label{fig7}
		The evolution of the  total loss(a) and style loss(b) during  training (paremeter $\alpha$). } 
\end{figure}

\begin{figure}
\begin{tabular}{c}
\centering
			\includegraphics[width=0.9\textwidth]{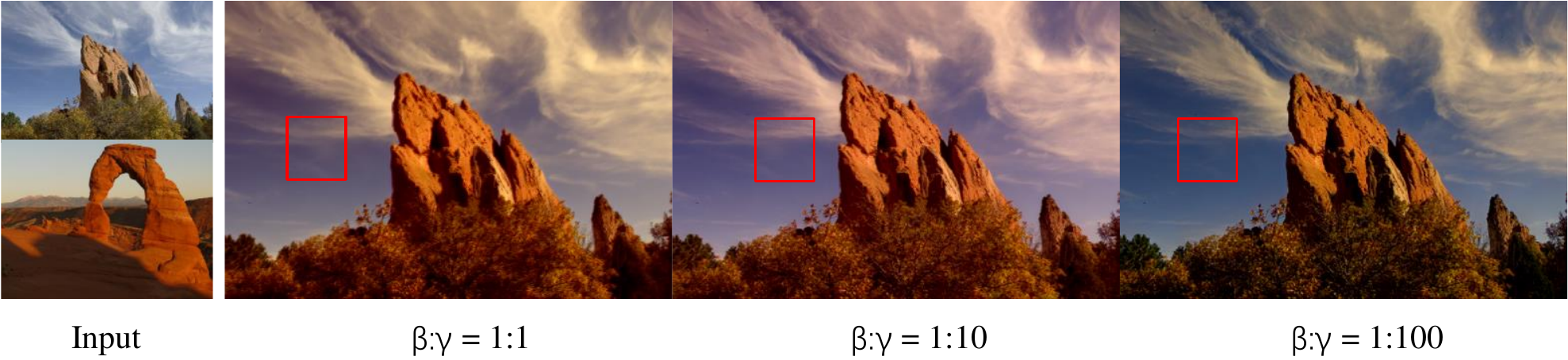}
\end{tabular}
	\caption 
	{ \label{fig8}
		The photorealistic stylization results for  different parameter values $\beta$:$\gamma$. (input pair (top: content, bottom: style)) } 
\end{figure}

\textit{2)\textbf{ Parameter($\alpha$) in Loss Function:}} The parameter $\alpha$ controls the weight of the style loss. It is set to 0.2, 0.5, 0.8. Other parameters are set as follows: $\beta$:$\gamma$ is 1:1 ($\beta$=1, $\gamma$=1), MFA is feature aggregation strategy. In Fig. \ref{fig6}, we show a set of images to compare the results obtained with different values of parameter $\alpha$. The plots of the total loss and style loss are shown in the Fig. \ref{fig7}. The results of the two quantitative evaluation criteria are summarized in Table \ref{table1}.

Fig. \ref{fig6} shows the results obtained with different values of parameter $\alpha$. As $\alpha$ increases, the degree of stylization increases visibly. The details in the red box are completely preserved, and its style is changed. In Fig. \ref{fig7}, the network quickly converges no matter what value the parameter $\alpha$ takes at  200 iterations. As the parameters $\alpha$ increases, the speed of the network convergence also increases. The final total loss converges to a close level. As for the style loss, the initial value is quite different for different values of parameter $\alpha$. But with fast convergence, the style loss values of becoming similar, regardless of the value of $\alpha$. This shows that no matter what the value of $\alpha$ is, our model is stable and convergent. In the test, the largest initial value shows the smallest loss of style. This shows that we can control the stylization of the generated image by adjusting the parameters $\alpha$. Since the content and style information are antagonistic, content information will be affected more when the weight of style loss is greater. As shown in Table \ref{table1}, when $\alpha$ is large, the style evaluation is excellent but the content index is poor. Based on these observations, the value of $\alpha$ is set to 0.8.

\begin{table}[!t] 
	\centering
	\caption{\label{table1}Quantitative evaluation of the impact of different parameter $\alpha$} 
	\setlength{\tabcolsep}{6mm}{
\begin{tabular}{cccc}
			\toprule 
			& 0.2 & 0.5 & 0.8 \\
			\midrule 
			Gram loss & 14.308 & 11.938 & \textbf{8.996} \\ 
			SSIM & \textbf{0.8056} & 0.663 & 0.612 \\ 
			\bottomrule 
\end{tabular}} 
\end{table}

\begin{figure}[!t]
\begin{center}
\begin{tabular}{c}
			\includegraphics[width=0.45\textwidth]{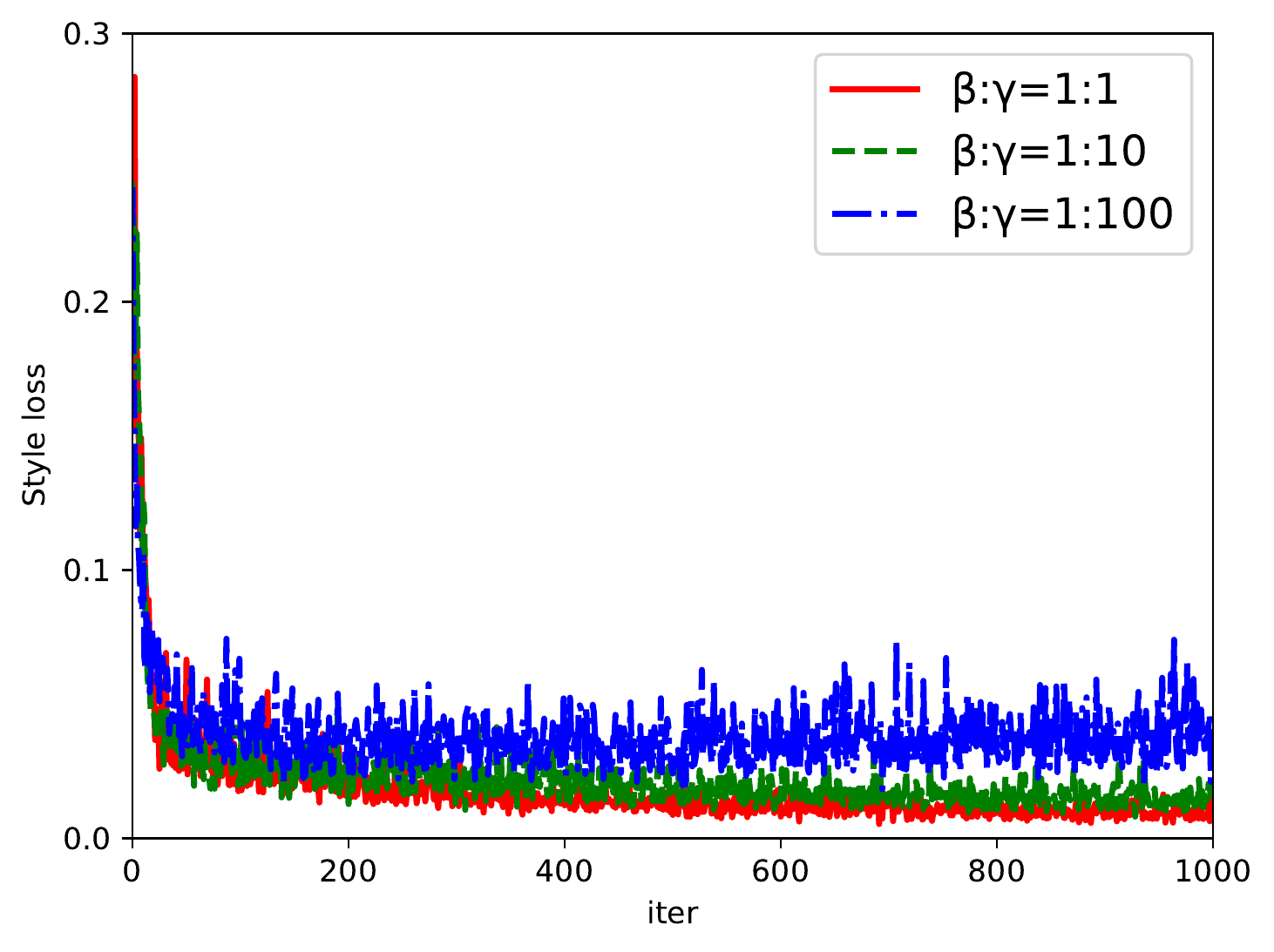}
\end{tabular}
\end{center}
	\caption 
	{ \label{fig9}
		The evolution of the style loss during  training (parameter $\beta$:$\gamma$).} 
\end{figure}

\textit{3)\textbf{ Parameters ($\beta$:$\gamma$) of the Loss Function:}} In the \cite{li2018densefuse}, Li et al. control the performance of  image fusion by the ratio of SSIM loss and content loss. We adopt the same strategy in our work. Accordingly, this parameter is set to 1:1, 1:10 and 1:100 ($\beta$=1, $\gamma$=1, 10, 100). Other parameters are set as follows: $\alpha$=0.8, MFA is used for feature aggregation. Since the change of this parameter will affect the value of the total loss, we only observe the change of style loss. Fig. \ref{fig8} shows the results obtained for different values of parameter $\beta$:$\gamma$. The plot of style loss is shown in the Fig. \ref{fig9}. Table \ref{table2} summarises the results measured using the  two quantitative evaluation criteria.

\begin{table}[!t] 
\centering
\caption{\label{table2}The impact of meta parameter $\beta$:$\gamma$} 
\setlength{\tabcolsep}{6mm}{
\begin{tabular}{cccc}
			\toprule 
			& 1:1 & 1:10 & 1:100 \\
			\midrule 
			Gram loss & \textbf{8.996} & 12.726 & 13.922 \\ 
			SSIM & 0.612 & 0.705 & \textbf{0.748} \\ 
			\bottomrule 
\end{tabular}} 
\end{table}

As shown in Fig. \ref{fig8}, the clouds in the red frame gradually become clear because of the increasing $\gamma$. In addition, the color distribution of the stylized image is closer to the content image, but it still retains some style features, such as red mountains. This proves that the change of parameter $\beta$:$\gamma$ can enhance the details of the content while retaining the style elements. In Fig. \ref{fig9}, the style loss drops rapidly by the first 200 iterations. When $\beta$:$\gamma$ is set to 1:1, the convergence of the style loss is the fastest. In addition, the blue line ($\beta$:$\gamma$=1:100) always exhibits large fluctuations, which are difficult to stabilize. The other two lines basically converge to the same level when the parameter $\beta$:$\gamma$ is set as 1:1 and 1:10. As shown in Table \ref{table2}, we can refine the image content by changing $\beta$:$\gamma$, but it will lead to a loss of style. Based on these results, we choose $\beta$:$\gamma$=1:1 ($\beta$=1, $\gamma$=1).

\begin{figure}
\begin{tabular}{c}
\centering
\includegraphics[width=0.85\textwidth]{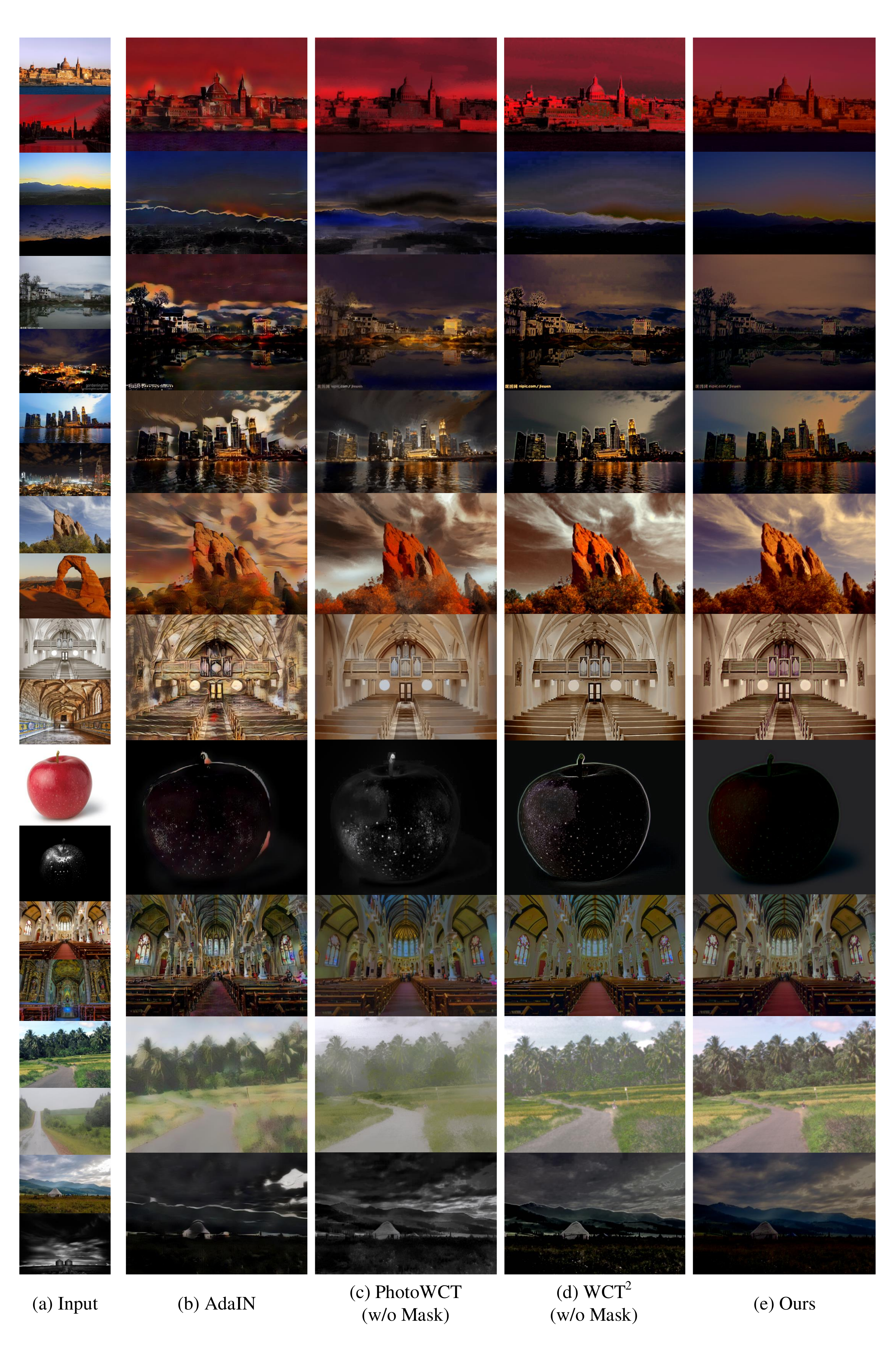}
\end{tabular}
\caption 
{ \label{fig10}
Photorealistic stylization results: (a) input pair (top: content, bottom: style); (b) AdaIN; (c) PhotoWCT (without mask); (d) WCT$^2$ (without mask); (e) Our method (UMFA). } 
\end{figure}
\subsection{Result Analysis}  
\subsubsection{Subjective evaluation}
In Fig. \ref{fig10}, we show the results obtained by existing methods and UMFA.  The results suggest that  the AdaIN\cite{huang2017arbitrary} method renders the highest degree of abstraction. PhotoWCT\cite{li2018closed} and WCT$^2$\cite{yoo2019photorealistic}, improved by whitening and coloring transforms (WCT),  retain the structural information of the image better. However,  the stylization effect of these two methods is unnatural. Besides, there are too many style elements that affect the realism of the photo. UMFA produces a natural stylization effect, closest to a real photo with minimum artifacts.

In order to better verify the effect of our method on the subjective experience of people, we also set up a comparison of stylized results produced by different methods. We scramble the pictures generated by different methods in Fig. \ref{fig10}, and ask 100 testers to choose the pictures that they think have the best photorealistic style transfer effect. As shown in Table \ref{table5}, the percentage indicates the proportion of the tester choosing the result of this method as the "Best stylization". Since they can only choose one method, the sum of the overall proportions is one. Among them, the proportion of choosing our method(UMFA) is the highest, that the result of UMFA is the most popular.
\begin{figure}
\begin{tabular}{c}
\centering
			\includegraphics[width=0.85\textwidth]{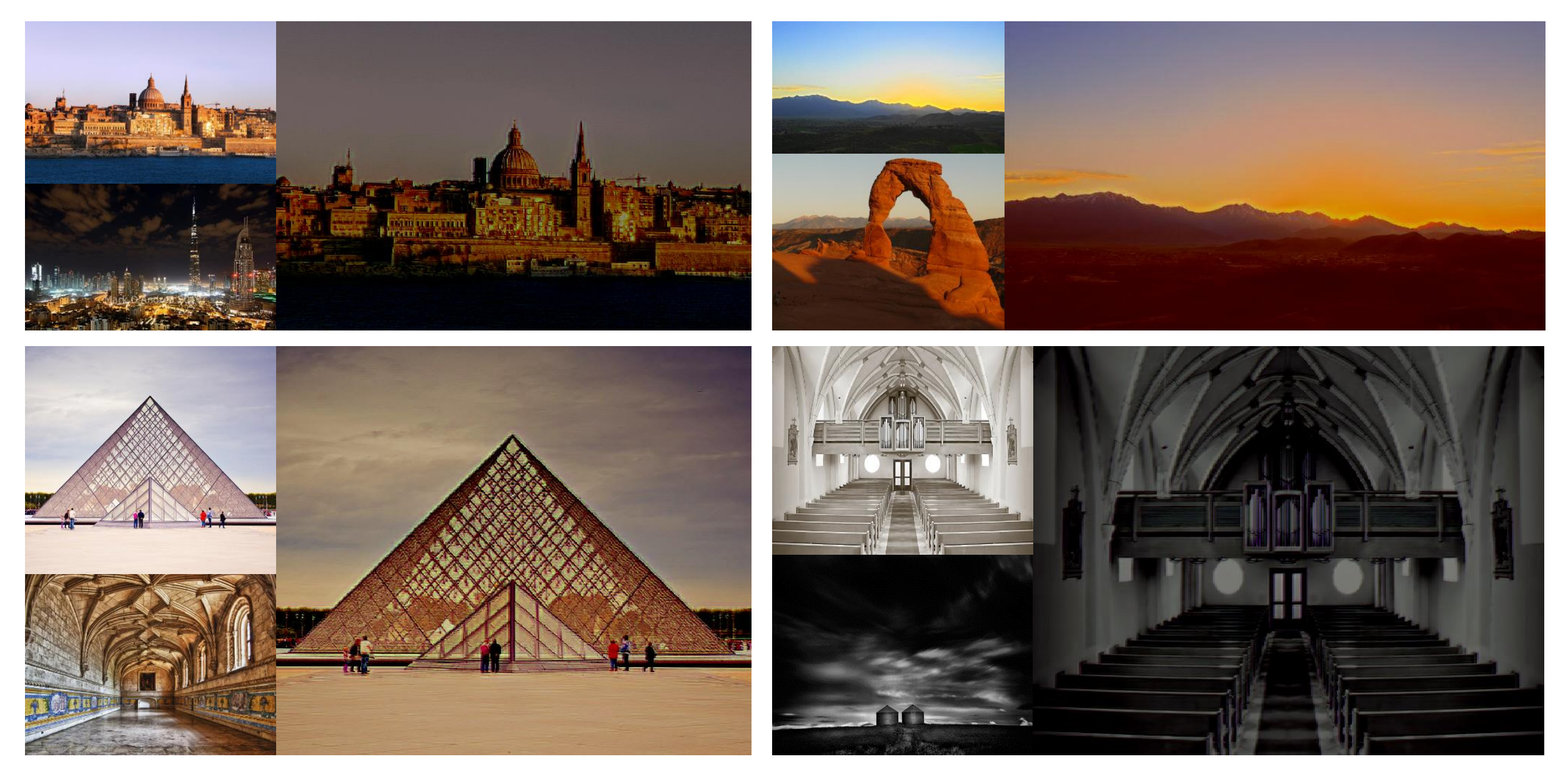}
\end{tabular}
\caption 
	{ \label{fig11}
	The result of using different structured content and style images.} 
\end{figure}

\begin{table}[!t] 
	\centering
	\caption{\label{table5}Subjective evaluation results of different methods} 
	\setlength{\tabcolsep}{3mm}{
\begin{tabular}{ccccc}
			\toprule 
			& AdaIN & PhotoWCT & WCT$^2$) & Ours(UMFA) \\
			&  & (w/o Mask) & (w/o Mask) &  \\
			\midrule 
			Best stylization & 13.2\% & 25.1\% & 24.8\% & \textbf{36.9\%} \\ 
			\bottomrule 
\end{tabular}} 
\end{table}

The main purpose of photorealistic style transfer is to change the style elements of the image without changing the main body of the image content. Therefore, in order to achieve a better stylization effect, users usually use pictures with similar structures when using algorithms. In addition to the similar image structure, we also performed photorealistic style transfer tests on other images to prove the effectiveness of our method. In Fig. \ref{fig11}, we show four sets of photorealistic stylized images, where the upper left corner of each part is the content image, the lower left corner is the style image, and the right side is the stylized image.

\subsubsection{Objective evaluation}
%In order to objectively evaluate the stylization ability of various methods, we use two evaluation indicators including Gram loss and SSIM. 
Table \ref{table6} presents the average values of the two objective criteria used for the evaluation of stylized images (Gram loss and SSIM). The best values are indicated in \textbf{bold}. As shown in Table \ref{table6}, UMFA has the smallest Gram loss and the largest SSIM which means that our photorealistic stylization result has a similar style to the style image and retains the structural information of the content image.

\begin{table}[htbp] 
	\centering
	\caption{\label{table6}Quantitative evaluation results of different methods } 
	\setlength{\tabcolsep}{3mm}{
\begin{tabular}{ccccc}
			\toprule 
			& AdaIN & PhotoWCT & WCT$^2$) & Ours(UMFA) \\
			&  & (w/o Mask) & (w/o Mask) &  \\
			\midrule 
			Gram loss & 11.356 & 11.367 & 9.323 & \textbf{8.996} \\ 
			SSIM & 0.347 & 0.423 & 0.516 & \textbf{0.612} \\ 
			\bottomrule 
\end{tabular}} 
\end{table}

\subsection{Runtime} 
Table \ref{table4} shows the runtime of AdaIN\cite{huang2017arbitrary}, PhotoWCT\cite{li2018closed}, WCT$^2$\cite{yoo2019photorealistic} and our method(UMFA). The numbers in the table represent the time required for various methods to generate a stylized picture at different image sizes. The unit of the number is seconds. The smallest values are indicated in \textbf{bold} which means fastest running speed. The results of our method are denoted in \textcolor{red}{red}. For PhotoWCT and WCT$^2$, we do not use mask to compare various methods at the same level. The post-processing time of the PhotoWCT method is too long. We use the time of stylization process as the comparison objective. AdaIn is the fastest method because it is stylized by matching the mean and variance requiring the small amount of calculation. In addition, it only uses the deepest features for style transfer, reducing the amount of calculation. But for the same reason, the stylization effect of the AdaIN method is the worst. PhotoWCT and WCT$^2$ take a similar amount of  time because they are based on the same method, but PhotoWCT needs more processing time to achieve the effect closed to WCT$^2$. Compared with the Adain, the method based on whitening and coloring transform requires a larger amount of calculation. Other operations introduced in the encoder and decoder process further increase the runtime. Among the four methods, UMFA does not take the least time. However, it can achieve real-time photorealistic style transfer and is an order of magnitude faster than the other two popular methods. Inspired by AdaIN, we changed the network structure, added MFA blocks and perform multi-layer stylization operations. As a result, we have achieved a balance between photorealistic stylization and time consumption.
\begin{table}[htbp] 
	\centering
	\caption{\label{table4}A comparison of runtime for different methods(Time/sec)}
	\setlength{\tabcolsep}{3mm}{
\begin{tabular}{ccccc}
			\toprule 
			Iamge Size & AdaIN & PhotoWCT & WCT$^2$ & Ours(UMFA) \\
			\midrule 
			256 $\times$ 256 & \textbf{0.046} & 3.44 & 5.04 & \textcolor{red}{0.14} \\ 
			512 $\times$ 512 & \textbf{0.12} & 3.48 & 5.64 &\textcolor{red}{0.24} \\ 
			1024 $\times$ 1024 & \textbf{0.35} & 3.66 & 6.06 & \textcolor{red}{0.64}\\  
			\bottomrule 
\end{tabular}} 
\end{table}

\section{CONCLUSION}
\label{sec5}
In this paper, we propose a photorealistic style transfer model named UMFA based on U-Net and multi-layer feature aggregation. 
First, the encoder, which uses the dense block in the downsampling module, can make full use of the image information to capture details. 
The "transfer module" is designed to enrich multi-scale features and enhance the expressiveness of stylization results by AdaIN and MFA blocks. 
The upsampling and skip connection of the U-Net structure are designed for the decoder to reconstruct and output the stylized image. 
The experimental results show that the UMFA method enhances the content detail in the stylized image and retains the texture of the photo, achieving style transfer with a natural photorealistic effect.

\section{Acknowledgments}
This work was supported by the National Key Research and Development Program of China (Grant No. 2017YFC1601800), the National Natural Science Foundation of China (61672265, U1836218), the 111 Project of Ministry of Education of China (B12018), and the UK EPSRC (EP/N007743/1, MURI/EPSRC/DSTL, EP/R018456/1).

%%%%% References %%%%%

\bibliography{report}   % bibliography data in report.bib

\begin{thebibliography}{10}

\bibitem{laffont2014transient}
P.-Y. Laffont, Z.~Ren, X.~Tao, {\em et~al.}, ``Transient attributes for
  high-level understanding and editing of outdoor scenes,'' {\em ACM
  Transactions on graphics (TOG)} {\bf 33}(4), 1--11  (2014).

\bibitem{luan2017deep}
F.~Luan, S.~Paris, E.~Shechtman, {\em et~al.}, ``Deep photo style transfer,''
  in {\em Proceedings of the IEEE Conference on Computer Vision and Pattern
  Recognition},  4990--4998  (2017).

\bibitem{akl2018survey}
A.~Akl, C.~Yaacoub, M.~Donias, {\em et~al.}, ``A survey of exemplar-based
  texture synthesis methods,'' {\em Computer Vision and Image Understanding}
  {\bf 172}, 12--24  (2018).

\bibitem{efros2001image}
A.~A. Efros and W.~T. Freeman, ``Image quilting for texture synthesis and
  transfer,'' in {\em Proceedings of the 28th annual conference on Computer
  graphics and interactive techniques},  341--346  (2001).

\bibitem{song2016sufficient}
Z.-C. Song and S.-G. Liu, ``Sufficient image appearance transfer combining
  color and texture,'' {\em IEEE Transactions on Multimedia} {\bf 19}(4),
  702--711  (2016).

\bibitem{gatys2016image}
L.~A. Gatys, A.~S. Ecker, and M.~Bethge, ``Image style transfer using
  convolutional neural networks,'' in {\em Proceedings of the IEEE conference
  on computer vision and pattern recognition},  2414--2423  (2016).

\bibitem{zhang2018real}
X.~Zhang, Y.~Luan, and X.~Li, ``Real-time image style transformation based on
  deep learning,'' {\em Journal of Electronic Imaging} {\bf 27}(4), 043045
  (2018).

\bibitem{zhang2018multi}
H.~Zhang and K.~Dana, ``Multi-style generative network for real-time
  transfer,'' in {\em Proceedings of the European Conference on Computer Vision
  (ECCV)},  0--0  (2018).

\bibitem{krizhevsky2012imagenet}
A.~Krizhevsky, I.~Sutskever, and G.~E. Hinton, ``Imagenet classification with
  deep convolutional neural networks,'' in {\em Advances in neural information
  processing systems},  1097--1105  (2012).

\bibitem{jing2019neural}
Y.~Jing, Y.~Yang, Z.~Feng, {\em et~al.}, ``Neural style transfer: A review,''
  {\em IEEE transactions on visualization and computer graphics}   (2019).

\bibitem{gatys2017controlling}
L.~A. Gatys, A.~S. Ecker, M.~Bethge, {\em et~al.}, ``Controlling perceptual
  factors in neural style transfer,'' in {\em Proceedings of the IEEE
  Conference on Computer Vision and Pattern Recognition},  3985--3993  (2017).

\bibitem{gu2018arbitrary}
S.~Gu, C.~Chen, J.~Liao, {\em et~al.}, ``Arbitrary style transfer with deep
  feature reshuffle,'' in {\em Proceedings of the IEEE Conference on Computer
  Vision and Pattern Recognition},  8222--8231  (2018).

\bibitem{cheng2019structure}
M.-M. Cheng, X.-C. Liu, J.~Wang, {\em et~al.}, ``Structure-preserving neural
  style transfer,'' {\em IEEE Transactions on Image Processing} {\bf 29},
  909--920  (2019).

\bibitem{hu2020aesthetic}
Z.~Hu, J.~Jia, B.~Liu, {\em et~al.}, ``Aesthetic-aware image style transfer,''
  in {\em Proceedings of the 28th ACM International Conference on Multimedia},
  3320--3329  (2020).

\bibitem{texler2020arbitrary}
O.~Texler, D.~Futschik, J.~Fi{\v{s}}er, {\em et~al.}, ``Arbitrary style
  transfer using neurally-guided patch-based synthesis,'' {\em Computers \&
  Graphics} {\bf 87}, 62--71  (2020).

\bibitem{li2019high}
M.~Li, C.~Ye, and W.~Li, ``High-resolution network for photorealistic style
  transfer,'' {\em arXiv preprint arXiv:1904.11617}   (2019).

\bibitem{yang2020te141k}
S.~Yang, W.~Wang, and J.~Liu, ``Te141k: Artistic text benchmark for text effect
  transfer,'' {\em IEEE Transactions on Pattern Analysis and Machine
  Intelligence}   (2020).

\bibitem{li2019chinese}
G.~Li, J.~Zhang, D.~Chen, {\em et~al.}, ``Chinese flower-bird character
  generation based on pencil drawings or brush drawings,'' {\em Journal of
  Electronic Imaging} {\bf 28}(3), 033029  (2019).

\bibitem{wang2020photographic}
L.~Wang, Z.~Wang, X.~Yang, {\em et~al.}, ``Photographic style transfer,'' {\em
  The Visual Computer} {\bf 36}(2), 317--331  (2020).

\bibitem{yoo2019photorealistic}
J.~Yoo, Y.~Uh, S.~Chun, {\em et~al.}, ``Photorealistic style transfer via
  wavelet transforms,'' in {\em Proceedings of the IEEE International
  Conference on Computer Vision},  9036--9045  (2019).

\bibitem{sunkavalli2010multi}
K.~Sunkavalli, M.~K. Johnson, W.~Matusik, {\em et~al.}, ``Multi-scale image
  harmonization,'' {\em ACM Transactions on Graphics (TOG)} {\bf 29}(4), 1--10
  (2010).

\bibitem{pitie2005n}
F.~Pitie, A.~C. Kokaram, and R.~Dahyot, ``N-dimensional probability density
  function transfer and its application to color transfer,'' in {\em Tenth IEEE
  International Conference on Computer Vision (ICCV'05) Volume 1},   {\bf 2},
  1434--1439, IEEE  (2005).

\bibitem{yoo2013local}
J.-D. Yoo, M.-K. Park, J.-H. Cho, {\em et~al.}, ``Local color transfer between
  images using dominant colors,'' {\em Journal of Electronic Imaging} {\bf
  22}(3), 033003  (2013).

\bibitem{levin2007closed}
A.~Levin, D.~Lischinski, and Y.~Weiss, ``A closed-form solution to natural
  image matting,'' {\em IEEE transactions on pattern analysis and machine
  intelligence} {\bf 30}(2), 228--242  (2007).

\bibitem{li2018closed}
Y.~Li, M.-Y. Liu, X.~Li, {\em et~al.}, ``A closed-form solution to
  photorealistic image stylization,'' in {\em Proceedings of the European
  Conference on Computer Vision (ECCV)},  453--468  (2018).

\bibitem{an2020ultrafast}
J.~An, H.~Xiong, J.~Huan, {\em et~al.}, ``Ultrafast photorealistic style
  transfer via neural architecture search.,'' in {\em AAAI},  10443--10450
  (2020).

\bibitem{huang2017arbitrary}
X.~Huang and S.~Belongie, ``Arbitrary style transfer in real-time with adaptive
  instance normalization,'' in {\em Proceedings of the IEEE International
  Conference on Computer Vision},  1501--1510  (2017).

\bibitem{ronneberger2015u}
O.~Ronneberger, P.~Fischer, and T.~Brox, ``U-net: Convolutional networks for
  biomedical image segmentation,'' in {\em International Conference on Medical
  image computing and computer-assisted intervention},  234--241, Springer
  (2015).

\bibitem{yu2018deep}
F.~Yu, D.~Wang, E.~Shelhamer, {\em et~al.}, ``Deep layer aggregation,'' in {\em
  Proceedings of the IEEE conference on computer vision and pattern
  recognition},  2403--2412  (2018).

\bibitem{vgg}
K.~Simonyan and A.~Zisserman, ``Very deep convolutional networks for
  large-scale image recognition,'' in {\em International Conference on Learning
  Representations},   (2015).

\bibitem{johnson2016perceptual}
J.~Johnson, A.~Alahi, and L.~Fei-Fei, ``Perceptual losses for real-time style
  transfer and super-resolution,'' in {\em European conference on computer
  vision},  694--711, Springer  (2016).

\bibitem{dumoulin2016learned}
V.~Dumoulin, J.~Shlens, and M.~Kudlur, ``A learned representation for artistic
  style,'' {\em arXiv preprint arXiv:1610.07629}   (2016).

\bibitem{ulyanov2017improved}
D.~Ulyanov, A.~Vedaldi, and V.~Lempitsky, ``Improved texture networks:
  Maximizing quality and diversity in feed-forward stylization and texture
  synthesis,'' in {\em Proceedings of the IEEE Conference on Computer Vision
  and Pattern Recognition},  6924--6932  (2017).

\bibitem{li2017universal}
Y.~Li, C.~Fang, J.~Yang, {\em et~al.}, ``Universal style transfer via feature
  transforms,'' in {\em Advances in neural information processing systems},
  386--396  (2017).

\bibitem{nas}
B.~Zoph and Q.~Le, ``Neural architecture search with reinforcement learning,''
  in {\em International Conference on Learning Representations},   (2017).

\bibitem{isola2017image}
P.~Isola, J.-Y. Zhu, T.~Zhou, {\em et~al.}, ``Image-to-image translation with
  conditional adversarial networks,'' in {\em Proceedings of the IEEE
  conference on computer vision and pattern recognition},  1125--1134  (2017).

\bibitem{cao2020auto}
B.~Cao, H.~Zhang, N.~Wang, {\em et~al.}, ``Auto-gan: self-supervised
  collaborative learning for medical image synthesis,'' in {\em Proceedings of
  the AAAI Conference on Artificial Intelligence},   {\bf 34}(07), 10486--10493
   (2020).

\bibitem{zhou2019branchgan}
Y.-F. Zhou, R.-H. Jiang, X.~Wu, {\em et~al.}, ``Branchgan: Unsupervised mutual
  image-to-image transfer with a single encoder and dual decoders,'' {\em IEEE
  Transactions on Multimedia} {\bf 21}(12), 3136--3149  (2019).

\bibitem{zhu2017unpaired}
J.-Y. Zhu, T.~Park, P.~Isola, {\em et~al.}, ``Unpaired image-to-image
  translation using cycle-consistent adversarial networks,'' in {\em
  Proceedings of the IEEE international conference on computer vision},
  2223--2232  (2017).

\bibitem{liu2017image}
D.~Liu, B.~Wen, X.~Liu, {\em et~al.}, ``When image denoising meets high-level
  vision tasks: A deep learning approach,'' {\em arXiv preprint
  arXiv:1706.04284}   (2017).

\bibitem{he2016identity}
K.~He, X.~Zhang, S.~Ren, {\em et~al.}, ``Identity mappings in deep residual
  networks,'' in {\em European conference on computer vision},  630--645,
  Springer  (2016).

\bibitem{he2016deep}
K.~He, X.~Zhang, S.~Ren, {\em et~al.}, ``Deep residual learning for image
  recognition,'' in {\em Proceedings of the IEEE conference on computer vision
  and pattern recognition},  770--778  (2016).

\bibitem{huang2017densely}
G.~Huang, Z.~Liu, L.~Van Der~Maaten, {\em et~al.}, ``Densely connected
  convolutional networks,'' in {\em Proceedings of the IEEE conference on
  computer vision and pattern recognition},  4700--4708  (2017).

\bibitem{li2018densefuse}
H.~Li and X.-J. Wu, ``Densefuse: A fusion approach to infrared and visible
  images,'' {\em IEEE Transactions on Image Processing} {\bf 28}(5), 2614--2623
   (2018).

\bibitem{wang2004image}
Z.~Wang, A.~C. Bovik, H.~R. Sheikh, {\em et~al.}, ``Image quality assessment:
  from error visibility to structural similarity,'' {\em IEEE transactions on
  image processing} {\bf 13}(4), 600--612  (2004).

\bibitem{lin2014microsoft}
T.-Y. Lin, M.~Maire, S.~Belongie, {\em et~al.}, ``Microsoft coco: Common
  objects in context,'' in {\em European conference on computer vision},
  740--755, Springer  (2014).

\end{thebibliography}
\bibliographystyle{spiejour}   % makes bibtex use spiejour.bst

%%%%% Biographies of authors %%%%%

\vspace{2ex}\noindent\textbf{Dongyu Rao} received the B.E. degree in School of Metallurgy and Environment from Central South University, China, in 2019. He is a master student at the Jiangsu Provincial Engineering Laboratory of Pattern Recognition and Computational Intelligence, School of Artificial Intelligence and Computer Science, Jiangnan University. His research interests include style transfer, machine learning and deep learning.

\vspace{2ex}\noindent\textbf{Xiao-Jun Wu} received the Ph.D. degree in pattern recognition and intelligent system from the Nanjing University of Science and Technology, Nanjing, in 2002. He has been with the School of Information Engineering, Jiangnan University since 2006, where he is a Professor of pattern recognition and computational intelligence. He has published over 300 papers in his fields of research. His current research interests include pattern recognition, computer vision, and computational intelligence.

\vspace{2ex}\noindent\textbf{Hui Li} received the B.Sc. degree in School of Internet of Things Engineering from Jiangnan University, China, in 2015. He is currently a PhD student at the Jiangsu Provincial Engineerinig Laboratory of Pattern Recognition and Computational Intelligence, School of Artificial Intelligence and Computer Science, Jiangnan University. His research interests include image fusion, machine learning and deep learning.

\vspace{2ex}\noindent\textbf{Josef Kittler} received the B.A., Ph.D., and D.Sc. degrees from the University of Cambridge, in 1971, 1974, and 1991, respectively. He is a distinguished Professor of Machine Intelligence at the Centre for Vision, Speech and Signal Processing, University of Surrey, U.K. He conducts research in biometrics, video and image database retrieval, medical image analysis, and cognitive vision. He published the textbook Pattern Recognition: A Statistical Approach and over 700 scientific papers.

\vspace{2ex}\noindent\textbf{Tianyang Xu} received the B.Sc. degree in electronic science and engineering from Nanjing University, Nanjing, China, in 2011. He received the PhD degree at the School of Internet of Things Engineering, Jiangnan University, Wuxi, China, in 2019. He is currently a research fellow at the Centre for Vision, Speech and Signal Processing (CVSSP), University of Surrey, Guildford, United Kingdom. His research interests include visual tracking and deep learning.
\listoffigures
\listoftables

\end{spacing}
\end{document}